# Interpretable Inverse Design of Metal-Organic Frameworks with Large Language Model Agents

*Kyungmin Nam[1‡], Seunghee Han[1‡], and Jihan Kim[1*]*

1 Department of Chemical and Biomolecular Engineering, Korea Advanced Institute of Science and Technology, Daejeon 34141, Republic of Korea.

‡ These authors contributed equally

*Corresponding author: jihankim@kaist.ac.kr

# ABSTRACT

Inverse design of metal-organic frameworks (MOFs) requires navigating combinatorial spaces with costly property labels and opaque machine-learning models. We introduce LLM4MOF, a closed-loop multi-agent framework that converts a natural-language target into chemical hypotheses, constraints, diagnostic tests, and feedback. One agent proposes interpretable hypotheses over metal nodes, linkers, pore geometry, and functionality. Another converts them into constraints selecting MOFs defined by a node, linker, and topology. The Matchmaker forms four beams to attribute gains to geometry, chemistry, or metal choice: full hypothesis, chemistry, metal only, and random baseline. Blind to database landscapes, LLM4MOF enriches top performers across six adsorption, separation, and electronic-structure tasks within 400 evaluations. It also designs and live-simulates de novo MOFs spanning $H_2$ storage, $SF_6$ capture, and $C_2H_6/C_2H_4$ separation, deriving a distinct design rule for each objective. Under an identical nominal evaluation budget it consistently outperforms random search, Bayesian optimization, and genetic algorithms, and the outcome is insensitive to the language-model backend. All of this uses not a single property-labeled training structure, whereas generative alternatives train on thousands to hundreds of thousands. A new objective requires only a natural-language request: interpretable inverse design at a fraction of the data cost of existing approaches.

# INTRODUCTION

Inverse design seeks to identify or construct materials that satisfy target properties before undertaking costly synthesis, characterization, or high-fidelity simulation.[1,2,3] Conventional genetic algorithms and Bayesian optimization have been widely used to search candidate materials toward target properties,[4-6] and more recent generative models have expanded the ability to propose new candidates directly.[7-9] However, these approaches often require large property-labeled datasets, while reliable property labels are typically obtained from simulations or experiments.[10,11] In addition, although many machine-learning-based inverse-design workflows learn statistical associations between material structures and target properties,[12] these associations are not always directly interpretable as chemical design insight. As a result, they provide limited guidance on which chemical or structural features should be selected, modified, or avoided to satisfy a target objective.[13,14]

These limitations are amplified in metal-organic frameworks (MOFs), which span a vast combinatorial space of metal-node and linker combinations, while only a small fraction has reliable property labels.[5,15-17] Expanding these datasets is also costly because many MOFs contain large unit cells and structurally complex pore networks, making simulation or experimental characterization difficult to scale.[18,19] Genetic algorithms have been used to search MOF design spaces, but they often require many evaluated structures to train surrogate models and repeated simulation cycles to update the search.[20,21] Bayesian optimization is often most effective in low-dimensional continuous search spaces, but MOF applications require surrogate representations of discrete categorical building blocks such as topologies, metal nodes, and linkers, and provide limited chemically interpretable feedback.[22,23] Generative models can propose new structures by learning from existing MOF databases.[24,25] Yet they also depend on large training datasets, and the chemical basis for why a generated structure should satisfy a target objective is not always explicit.

To this end, large language models (LLMs) offer a promising route to address this bottleneck. Because LLMs are pretrained on broad corpora containing scientific and chemical knowledge, they can propose design hypotheses directly from a stated objective, without assembling a labeled property dataset or training a task-specific model.[26] Recent studies have shown that LLM agents can extract synthesis conditions from the literature, support experimental planning, optimize synthesis conditions, and enable autonomous research loops by connecting scientific knowledge with external tools and feedback.[27-33] LLMs have also been used for materials generation and modification, including the optimization of molecular or linker-level components.[34,35] However, directly generating MOFs from among billions of candidates is difficult for an LLM as it has not encountered most of these structures during pretraining, and it cannot reliably anticipate how a given metal node and linker combination will assemble into a framework with the desired pore geometry and shape. This motivates a framework in which LLMs provide chemically interpretable design reasoning, while structure construction, property evaluation, and feedback are handled by automated computational modules.[36,37]

Here, we introduce LLM4MOF, a closed-loop multi-agent framework that separates chemical reasoning from structure construction in MOF inverse design. Given a natural-language objective, LLM4MOF converts chemical hypotheses over nodes, linkers, topology, pore geometry, and functionality into executable constraints, evaluates matched or newly assembled MOFs, and returns interpretable feedback for the next design round. The loop runs autonomously without training a property-specific model for each objective. We validate LLM4MOF in two settings. In database mode, the agents remain blind to property values and global search-space information, yet enrich top-performing structures across six adsorption, separation, and electronic-structure tasks within roughly 400 evaluations. In discovery mode, the same loop proposes and simulates de novo MOFs under live simulation, adapting compact-micropore designs to conditions absent from the database. Together, these results establish an interpretable inverse-design strategy enabled by the modular node-linker-topology structure of MOFs.

# RESULTS

## LLM4MOF: a closed-loop multi-agent reasoning system

We developed LLM4MOF as a closed-loop framework that converts a natural-language MOF design objective into search and evaluation steps. As illustrated in **Figure 1**, each design task proceeds through five stages: (1) hypothesis generation, (2) constraint translation, (3) candidate matching, (4) hypothesis testing, and (5) feedback-driven refinement, with the full cycle repeated for ten iterations.

First, the hypothesis generator (Agent 1) proposes a structured design hypothesis from the target objective and the feedback from previous iterations, identifying candidate design principles at the level of metal nodes, linkers, pore geometry, and chemical functionality. Second, the constraint translator (Agent 2) converts this hypothesis into a constraint specification, including permitted metals, linker motifs, connectivity requirements, excluded groups, and geometric preferences. This separation preserves chemically interpretable design intent while reducing invalid or ambiguous constraints.

Third, the Matchmaker applies the translated constraints to the available candidate space and organizes the resulting candidates into four diagnostic beams: the full-hypothesis beam (Beam 1), the metal-linker chemistry beam (Beam 2), the metal-only beam (Beam 3), and a random baseline beam (Beam 4). These beams test whether performance improvements arise from the complete hypothesis, selected chemistry, metal node alone, or unconstrained sampling.

Fourth, the Hypothesis Testing module evaluates candidates in database or discovery mode. In database mode, the candidate properties are retrieved from the precomputed MOF property datasets, bypassing structure construction, so that the reasoning, translation, matching, and feedback loop can be validated without new simulations. In discovery mode, LLM4MOF proposes new MOFs, constructs their structures, and evaluates them via automated simulations, thereby testing whether the same loop can guide discovery when both structures and property values must be produced on demand.

Finally, the Feedback Generator compares the evaluated beams, identifies which chemical or geometric constraints are associated with successful candidates, and returns this evidence to Agent 1 as interpretable feedback. At the start of each new iteration, Agent 1 receives the most recent feedback in full, while earlier observations are summarized in a compact memory ledger. Through this process, LLM4MOF refines its hypotheses, relaxes overly restrictive constraints, and redirects the search toward promising MOF subspaces. Details of the diagnostic beams, execution modes, and implementation are provided in **Methods, Supplementary Notes S1-S3, Figure S1 and Figure S2**.

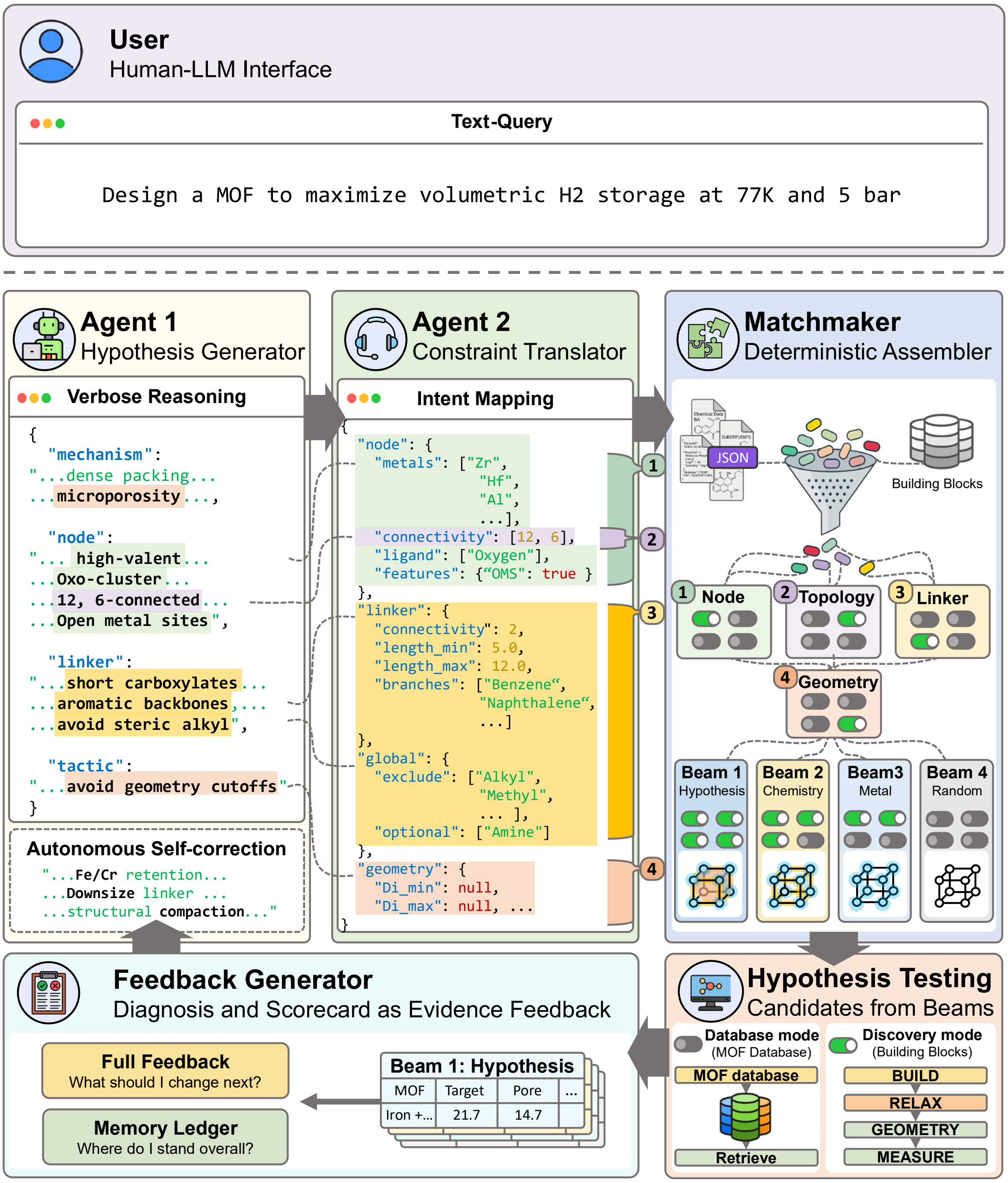

**Figure 1. Overview of the LLM4MOF closed-loop multi-agent workflow.** A user-defined target query is first converted by Agent 1 into a chemically interpretable hypothesis JSON, which Agent 2 then translates into a constraint JSON. The Matchmaker applies these constraints to the candidate space and organizes the matched candidates into four diagnostic beams (Beam 1, full hypothesis; Beam 2, metal-linker chemistry; Beam 3, metal only; Beam 4, random baseline). Candidate properties are obtained either in database mode using precomputed property values or in discovery mode using live simulations performed during the design loop. The Feedback Generator returns a feedback package to Agent 1 for the next iteration, consisting of the most recent full feedback report and a memory ledger of prior observations. The full closed loop is autonomously repeated for ten iterations.

## Database Mode: Rapid Enrichment toward Top-Performing MOFs

We first evaluated LLM4MOF in database mode, where precomputed property tables were used only as hidden evaluators rather than as searchable databases. The agents were not shown the structure list, database size, global property distribution, or best-performing structures, so this benchmark tested reasoning, translation, matching, and feedback rather than database screening. The benchmark covered six tasks from three databases: $H_2$ uptake from 14,173 PORMAKE-derived structures[25], $CH_4$ uptake, $CO_2$ uptake, and Xe/Kr selectivity from 51,145 hMOFs[17], and two band-gap objectives from 20,373 QMOFs[16]. To make these heterogeneous spaces searchable through Agent 2 constraints, the PORMAKE building blocks and database records were annotated with a shared chemistry-query metadata schema, so that hMOF and QMOF entries are filtered as whole structures while PORMAKE candidates are filtered as building-block combinations. Metadata details are provided in **Supplementary Note S4, Tables S3 and S4**.

Each task was run for ten iterations in five independent replicates, with up to ten candidates per beam, corresponding to at most 40 property evaluations per iteration and 400 per run. **Figure 2** reports the replicate-averaged beam median at each iteration, showing that the full-hypothesis beam rapidly shifted from the database median toward top-performing regions across the six tasks. On average, it crossed the top-10% reference within approximately five iterations or approached the top-1% reference by the final iteration for all tasks except the low-band-gap minimization task. Database reference statistics and final-iteration full-hypothesis values are summarized in **Table S1**.

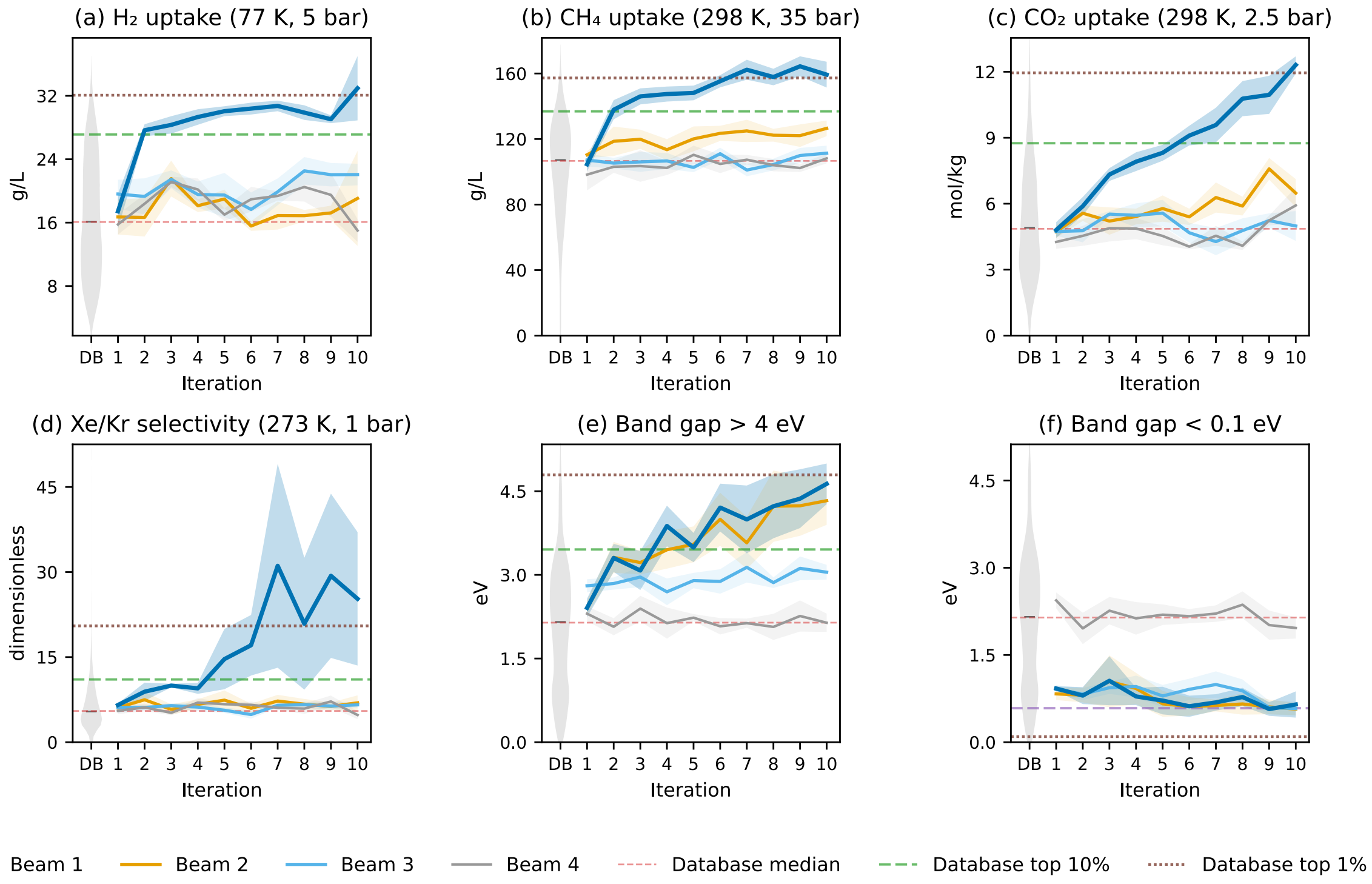

**Figure 2. Database-mode validation of LLM4MOF across six MOF design tasks.** Each panel shows the per-iteration median property value of candidates sampled from four diagnostic beams (Beam 1, full hypothesis; Beam 2, metal-linker chemistry; Beam 3, metal only; Beam 4, random baseline). Lines indicate the mean over five independent replicates, and shaded regions indicate ±1 SEM. The grey violin at DB shows the full database distribution, and horizontal reference lines mark the database median, top 10%, and top 1% values. For the low-band-gap task, lower values are better, so the top 10% and top 1% references correspond to the 10th and 1st percentiles, respectively.

Beam separation revealed which design axis drove enrichment, and this axis differed across properties rather than being fixed in advance. In the adsorption and selectivity tasks (**Figure 2a–d**), the full-hypothesis beam separated most clearly from the chemistry-only and metal-only beams, indicating that pore-geometry constraints were critical. For the large-band-gap objective (**Figure 2e**), the full-hypothesis and chemistry-only beams tracked closely, suggesting that chemical composition carried most of the design signal. For low-band-gap minimization (**Figure 2f**), no beam separated cleanly, indicating that this target is not governed by a single axis among geometry, chemistry, and metal choice.

To verify that the agent's success in Database Mode was driven by structural reasoning rather than stochastic chance, we analyzed the pooled hit rates of the retrieved candidates. Across the evaluated tasks, the hypothesis-constrained beam systematically enriched top-tier candidates relative to the baseline. For example, in the Xe/Kr selectivity task, 14.6% of the structures the agent proposed exceeded the database top-1% threshold, against 1.6% for the random sampling baseline. Enrichment was between four-fold and eighteen-fold on eight of the nine tasks, and larger on CH4 uptake, where the baseline returned one qualifying structure in five hundred. Comprehensive details about hit rate comparison are in **Table S2**.

Beyond locating top-performing regions, the loop also exposes the chemical logic behind each result. To see how enrichment emerged from the LLM-generated hypotheses, we traced representative hypothesis-to-structure trajectories for three of the tasks, one per database source (**Figure 3**). Each trajectory links the accumulated Agent 1 reasoning to a representative iteration-10 structure and to its motion through a task-relevant descriptor space. Each iteration-10 structure ranks within the top 0.01% of its database for the target property.

For $H_2$ uptake in the PORMAKE-derived database (**Figure 3a**), the trajectory moved from an initial open-metal-site, open-pore prior toward compact micropore confinement, consistent with the learned rule that volumetric uptake at 5 bar benefits from dense micropores rather than oversized pores. For $CO_2$ uptake at 298 K and 2.5 bar in the hMOF database (**Figure 3b**), the trajectory showed a feedback-driven correction: Agent 1 first emphasized open metal sites and polar functionality, a standard low-pressure $CO_2$ adsorption heuristic, but beam feedback redirected the hypothesis toward rigid aromatic pore walls, low substituent mass, and tight micropore confinement. The QMOF band-gap task (**Figure 3c**) instead followed an electronic-structure axis, moving toward closed-shell metal nodes and low-conjugation oxygen-donor linkers. Together, these trajectories show that LLM4MOF not only enriches high-performing candidates but also exposes and revises property-specific design logic. Additional trajectories for $CH_4$ uptake, Xe/Kr selectivity, and low-band-gap minimization are provided in **Figure S3**.

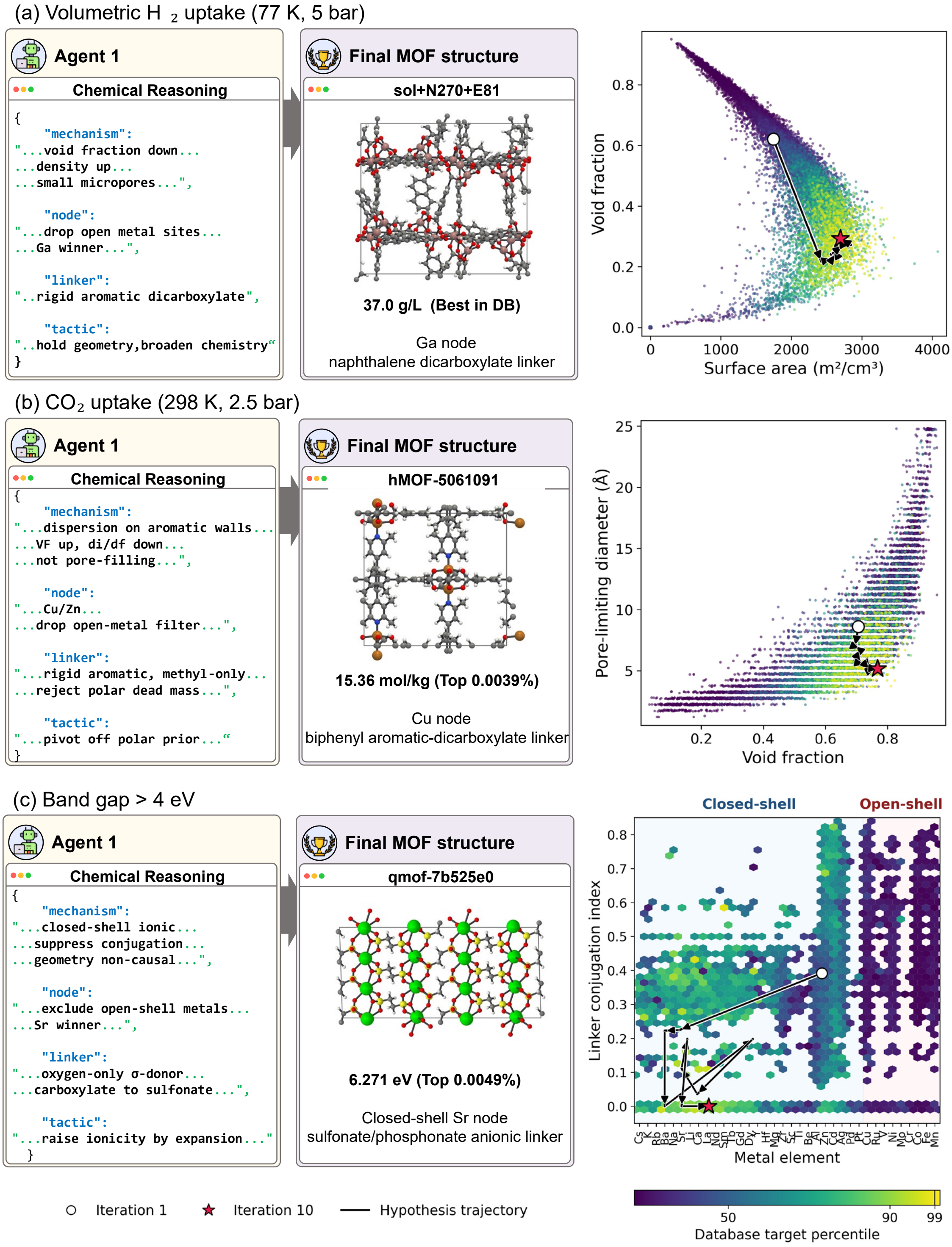

**Figure 3. Representative database-mode examples connecting Agent 1 reasoning to iteration-10 structures and descriptor-space trajectories**. Examples are shown for (a) $H_2$ uptake in the PORMAKE-derived database, (b) $CO_2$ uptake in the hMOF database, and (c) band-gap maximization in the QMOF database. Each row shows accumulated Agent 1 reasoning, a representative structure obtained at iteration 10, and the corresponding trajectory in a task-relevant database descriptor space. Atom colors in the structure panels are C, gray; O, red; H, white; N, blue; Ga, muted pink; Cu, bronze; Sr, green; and S, yellow. Background points are database structures colored by target-property percentile. White circles and red stars mark the mean of Beam 1 candidates sampled in iterations 1 and 10, respectively, and black arrows indicate the direction of hypothesis refinement.

The loop also adapts its hypotheses to how uptake is defined and to pressure. Applying the same protocol to $H_2$ uptake at 77 K measured both per unit volume and per unit mass, at 5 and 100 bar, we found that the learned design rules differ systematically across the four conditions. Volumetric uptake is governed by a constrained geometry window (i.e. dense micropores that pack capacity into a fixed volume), while gravimetric uptake instead favors low framework mass and high accessible pore volume. Pressure shifts both: from 5 to 100 bar, the preferred regions open up, mostly for gravimetric uptake consistent with the need for greater pore volume to store $H_2$ at high loading.

The Beam 1 and Beam 2 separation tracks this interpretation. For the volumetric uptake, Beam 1 separates clearly from Beam 2 since density, void fraction, and pore-size filtering are all needed to identify high-capacity structures. For gravimetric uptake at 5 bar, the two beams stay closer, indicating that light-framework chemistry alone already captures most of the signal. At 100 bar, however, Beam 1 again rises well above Beam 2, showing that chemistry is no longer sufficient and that explicit filtering toward extreme low density and high void fraction is required to reach the best high-pressure gravimetric structures. Across all four conditions, then, the loop adjusts both which design axis it relies on and how heavily, adapting its reasoning to the specific objective rather than applying a fixed rule. Detailed figures about $H_2$ uptake are presented in **Figures S4 and S5**.

## Robustness of Database-Mode Results to the Language-Model Backend

With all other components frozen, we re-ran the database-mode campaign with six different backends in the hypothesis-generating role, to test whether these results depend on the choice of language model. Each backend ran seven tasks in five replicates and was scored on its first attempt, so the benchmark measures deployable reliability rather than best-case capability as shown in **Table 1**.

The three frontier backends finish within 0.1 percentile points of one another, far inside the replicate-level spread of any one of them, so this benchmark does not resolve their ordering. Two mid-tier models follow at 86.7 and 86.6, and the smallest model reaches 79.5. The second mid-tier score carries a caveat, as that backend completed only 23 of 35 cells on the first attempt. Constraint strictness varied far more than the endpoint and is reported beside the score, because proposing narrow constraints is a stylistic property of a backend rather than an error. Above a capability threshold the backend is therefore interchangeable, and below it the failure appears in reliability before it appears in the score. The speed of enrichment is similarly stable. Every backend that completes its runs reaches the database top 10% in under five iterations on average. Per-task results, trajectories, and a control campaign under a second feedback sampler are provided in **Supplementary Note S13**, **Figures S8** and **S9**, and **Table S7**.

**Table 1. Six language-model backends in the hypothesis-generating role under a frozen configuration.** Scored on each backend's first attempt; 'no match' is constraint strictness, reported beside the score, not an error. The claude-sonnet-5 mean iterations-to-top-10% excludes one task it never reached. Final percentile is the final-iteration database percentile; completed cells are out of 35, from seven tasks with five replicates each; iteration counts are means over completed cells. Per-task scored replicate counts are smaller where a completed run's final iteration returned no matching candidates (Table S7).

| Model | Final percentile | Completed cells | Iterations to top 10% | No match | Unparseable |
|---|---|---|---|---|---|
| gpt-5.2 | 91.3 | 35 | 4.7 | 8.3% | 0% |
| gpt-5.6-sol | 91.3 | 34 | 3.7 | 24.6% | 0% |
| claude-opus-5 | 91.2 | 35 | 4.8 | 8.0% | 0% |
| claude-sonnet-5 | 86.7 | 35 | 4.2 | 2.3% | 0% |
| claude-haiku-4.5 | 86.6 | 23 | 4.7 | 11.8% | 7.1% |
| gpt-5.4-nano | 79.5 | 35 | 4.7 | 34.9% | 0% |

## Discovery Mode: Inverse Design of MOFs with Live Simulations

Having validated the closed loop in database mode, we next tested LLM4MOF in discovery mode, where no pre-existing MOF structures or precomputed property labels are available. Here, Agent 2 constraints define a candidate region in the combinatorial space of topologies, nodes, and linkers, and candidate MOFs are generated, structurally optimized, geometrically characterized, and evaluated by live simulations as the loop runs. This setting probes whether the same closed-loop reasoning can drive discovery when both the structures and their properties must be produced within the loop rather than retrieved.

The discovery-mode workflow is summarized in **Figures S1 and S2**. At each iteration, the Matchmaker generates candidate building-block combinations for the four diagnostic beams. For Beam 1, these combinations are first ranked by MOF2Zeo, a fast surrogate that we developed to predict pore geometry directly from a candidate's topology, node, and linker. This pre-ranking is worthwhile because pore geometry emerges only from the assembled framework, and not the individual building blocks, so identifying geometrically promising candidates would otherwise require assembling and relaxing every combination, which would be the most expensive step in the pipeline. MOF2Zeo lets Beam 1 concentrate this expense on the candidates most likely to satisfy the geometry hypothesis. The remaining beams skip this pre-ranking and pass directly to the structure generation. The code-level equivalence of the two operating modes is detailed in **Supplementary Note S1**.

Each iteration then proceeds through generation and simulation. To account for generation, relaxation, and geometry-matching failures, 100 candidates are generated for Beam 1 and 30 for each remaining beam; all are assembled with PORMAKE, relaxed with LAMMPS/UFF, and characterized with Zeo++. Beam 1 candidates advance according to post-relaxation Zeo++ geometry descriptors, and up to ten structures per beam are evaluated by RASPA3 GCMC simulations. Beam 1 candidates are selected by the geometry hypothesis, whereas candidates from the remaining beams are randomly selected from valid survivors, attributing Beam 1 gains to geometry-guided selection rather than sampling. Beam 1 also draws from a larger over-generated pool than the other beams, so the gap reflects selection intensity as well as hypothesis content; the database-mode beams carry no such asymmetry. Details of MOF2Zeo, its validation, and the full discovery-mode execution scheme are provided in **Supplementary Note S5** and **Figure S6**.

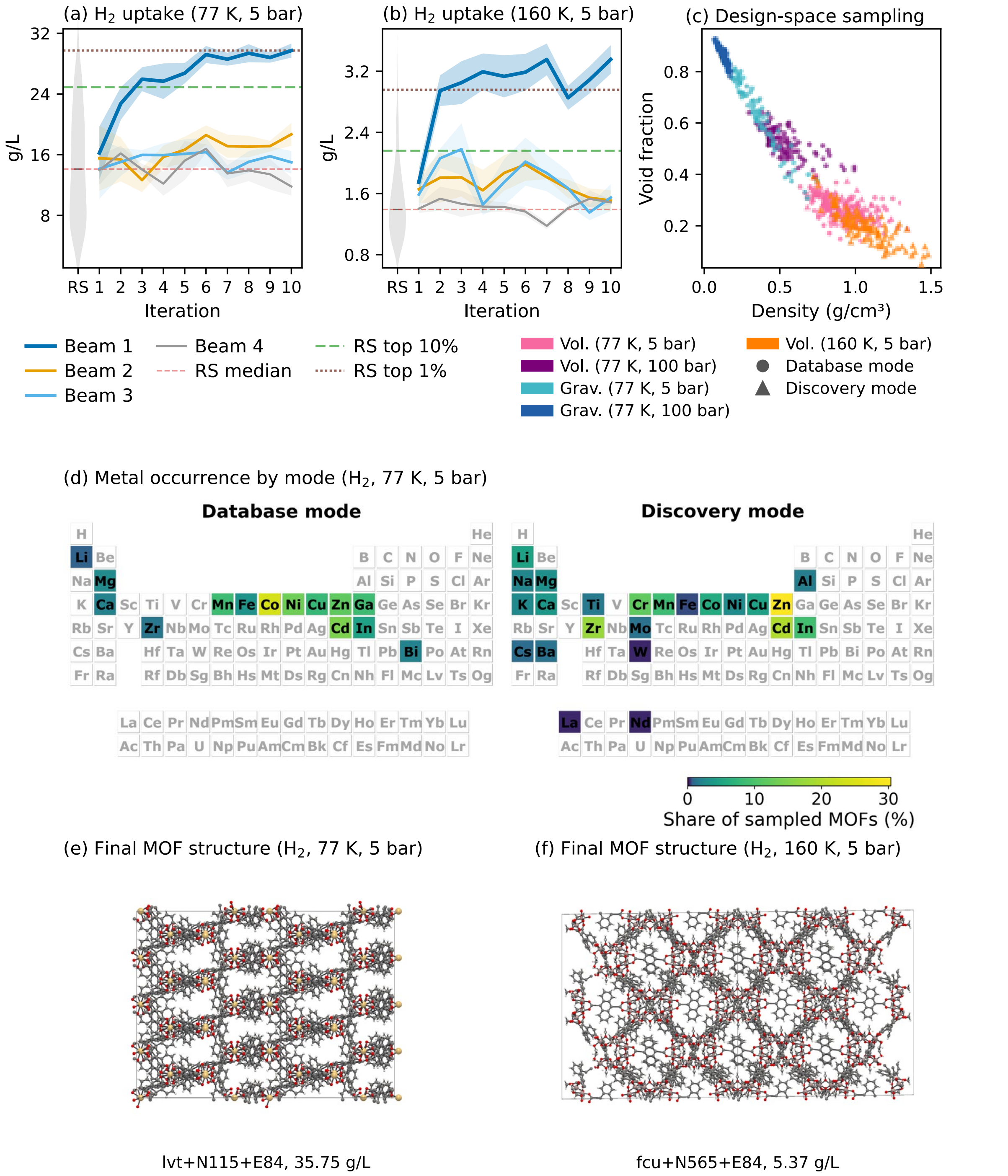

**Figure 4. Discovery-mode closed-loop search with live simulations.** (a,b) $H_2$ uptake at 5 bar under (a) 77 K and (b) 160 K. Lines show the mean beam median over five independent replicates, and shaded regions indicate ±1 SEM. Random-search references are computed from Beam 4 structures. (c) Density and void fraction of structures sampled by the hypothesis beam in database and discovery modes. Circles and triangles denote database-mode and discovery-mode structures, respectively. (d) Metal occurrence among sampled hypothesis-beam MOFs in database and discovery modes for the 77 K, 5 bar volumetric $H_2$ task. Colors indicate the fraction of sampled MOFs containing each metal element within each mode. (e) Structural view of the top-performing framework identified in the $H_2$ uptake live simulation at 77 K, 5 bar. (f) Structural view of the top-performing framework identified in the $H_2$ uptake live simulation at 160 K, 5 bar.

We then ran LLM4MOF in discovery mode for $H_2$ adsorption at 77 K and 160 K, both at 5 bar, to test whether the loop identifies high-performing structures under live simulation and adapts to temperature. Because discovery mode lacks an enumerated property database, randomly generated Beam 4 structures defined the reference distribution for the accessible generative space. At 77 K, the full-hypothesis beam exceeded the random-search top-10% reference and approached the top-1% reference within ten iterations (**Figure 4a)** with the best structure found by iteration 10 reaching 35.75 g/L (**Figure 4e**). At 160 K, the full-hypothesis beam again outperformed the chemistry-only, metal-only, and random beams, exceeding the random-search reference levels **(Figure 4b**). The best iteration-10 structure was an Al(III)-fcu framework with anthracene dicarboxylate linkers, reaching 5.37 g/L, consistent with the hypothesis that stronger $H_2$ confinement requires rigid, polarizable aromatic pore surfaces (**Figure 4f**). The shared anthracene dicarboxylate linker across different metals and topologies suggests that the geometry constraints repeatedly favored rigid aromatic shapes capable of forming favorable micropore environments for $H_2$ physisorption. The separation between Beam 1 and Beam 2 further shows that geometry selection, not chemistry alone, is essential for live discovery.

We next examined density and void fraction, descriptors repeatedly emphasized by Agent 1 for volumetric versus gravimetric $H_2$ storage. In this space, database-mode Beam 1 samples for gravimetric uptake occupy low-density, high-void-fraction regions, whereas volumetric samples shift toward higher density and lower-to-moderate void fraction (**Figure 4c**). Discovery candidates for volumetric $H_2$ uptake at 77 K and 5 bar fall in the same region, showing that the compact-micropore rule emerges consistently in both modes. At 160 K, candidates shift further toward denser, lower-void-fraction structures, consistent with stronger confinement when $H_2$ adsorption weakens at higher temperature.[38]

Finally, metal-occurrence maps show that discovery mode explores a broader node-chemistry space than database mode while retaining the geometry-controlled volumetric design regime (**Figure 4d**). Database-mode sampling concentrates in frequently retrieved metals, whereas discovery mode spans a wider range of nodes, reflecting the larger constructible space accessed by on-demand generation. Thus, both modes recover not a single metal-linker identity but a geometry-centered design principle realized through multiple chemical routes.

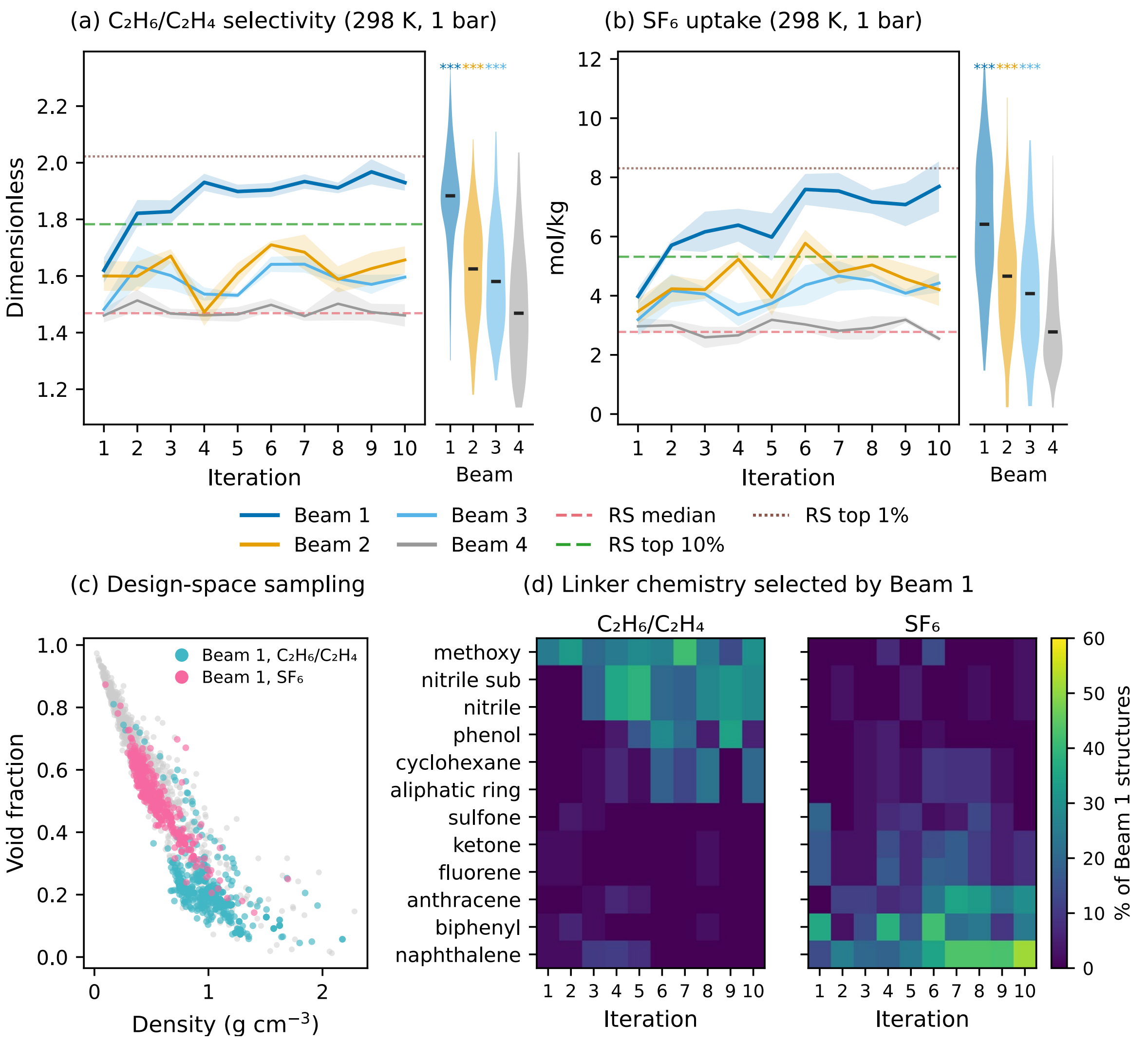

**Figure 5. Discovery-mode transfer to additional chemistries.** Closed-loop discovery, run unchanged, for (a) ethane-selective $C_2H_6/C_2H_4$ separation and (b) $SF_6$ capture, both at 298 K and 1 bar. Lines show the mean beam median over five independent replicates, and shaded regions indicate ±1 SEM. Dashed lines mark the median, top 10% and top 1% of the pooled random baseline (Beam 4). Violins show each beam's pooled distribution over all iterations and replicates; asterisks give one-sided Mann–Whitney significance against the random baseline (*** $p < 10{-}3$), computed on pooled structures. At the replicate level the full-hypothesis beam exceeds the random beam in five of five replicates in both tasks (one-sided sign test, $p = 0.031$). (c) Density and void fraction of structures sampled by the full-hypothesis beam in each task, over the pooled random-baseline cloud. (d) Fraction of full-hypothesis-beam structures per iteration whose linkers carry each functional-group tag, on rows shared between the two tasks. Screening thresholds and plotting conventions in **Supplementary Note S12**.

To test whether the closed loop transfers beyond hydrogen, we ran discovery mode unchanged on two objectives with different physics, ethane-selective $C_2H_6/C_2H_4$ separation and $SF_6$ capture, both at 298 K and 1 bar. Paraffin-selective adsorbents deliver polymer-grade ethylene in a single step, and computational screens place their best candidates near a selectivity of two[21,39]; $SF_6$ capture is likewise an active screening target.[40] In both tasks the replicate-averaged full-hypothesis beam crossed the random-search top-10% reference by the second iteration, individual replicates between the second and the sixth, and it exceeded the random beam in every replicate, ending at a median selectivity of 1.93 against 1.47 and a median $SF_6$ uptake of 7.7 mol $kg^{-1}$ against 2.8 (Figure 5). The two tasks resolved to opposite design logics (**Figure 5c,d**). For ethane selectivity, every replicate converged on dense, tight frameworks whose aliphatic pore walls maximize dispersion contact with ethane while offering no π or open-metal sites that would favor ethylene. The force field used here does not represent the π-complexation that drives ethylene selectivity in real materials, so the recovered rule is one of dispersion and confinement within that description (Note S12). For $SF_6$, the loop instead retained moderate micropores and did not converge on a common metal, delivering its five best structures with five different metals. The same loop thus learns a confinement-and-chemistry rule where selectivity demands one, and a geometry-centered rule where capacity does.

## Efficiency and Comparison with Inverse-Design Approaches

After confirming that discovery mode can identify high-performing MOFs through live simulation, we next asked whether the loop reaches this performance under the tight evaluation budget that live simulation imposes, where each evaluated structure carries the full cost of assembly, relaxation, and GCMC. We compared LLM4MOF against random sampling, a surrogate-assisted genetic algorithm[5], and Bayesian optimization[23] under an identical budget, with 40 evaluated structures per iteration over ten iterations, with each method repeated across five replicates. To match the iterative setting, the genetic algorithm and Bayesian optimization were initialized with 40 randomly evaluated structures at the first iteration and then proposed 40 additional structures per iteration using the accumulated evaluation history. Details of the genetic algorithm and Bayesian optimization are provided in **Supplementary Notes S6 and S7**.

LLM4MOF outperformed all three baselines (**Figure 6a**) in the live $H_2$ uptake task at 77 K and 5 bar. By iteration 10, it reached a median uptake of 29.3 g/L, compared with 21.8 g/L for the genetic algorithm, 17.7 g/L for the Bayesian optimization, and 14.3 g/L for random sampling. Although genetic algorithms and Bayesian optimization are powerful search strategies for materials design, both are constrained in this low-budget live-simulation setting. Genetic algorithms typically benefit from large evaluated populations and multiple optimization cycles, whereas Bayesian optimization must build an effective surrogate from a small number of sparse evaluations in a mixed categorical MOF design space. Their performance in this live benchmark was therefore limited by the small number of structures that could be evaluated.[41,42]

The LLM inference cost was small. Across five replicate runs, each ten-iteration campaign used 380 RASPA simulations, approximately 0.57 million GPT-5.2 tokens, and \$1.06 in LLM inference cost. The simulation count remained nearly constant across iterations, whereas token use and LLM cost increased gradually as the multi-turn context grew. The memory ledger limited this growth by carrying forward earlier evidence as a bounded summary rather than repeatedly expanding the prompt with all prior feedback. These results show that LLM4MOF can outperform random sampling, Bayesian optimization, and a data-limited genetic algorithm with a modest simulation budget and negligible LLM inference cost. Details of the LLM inference cost are provided in **Supplementary Note S8**.

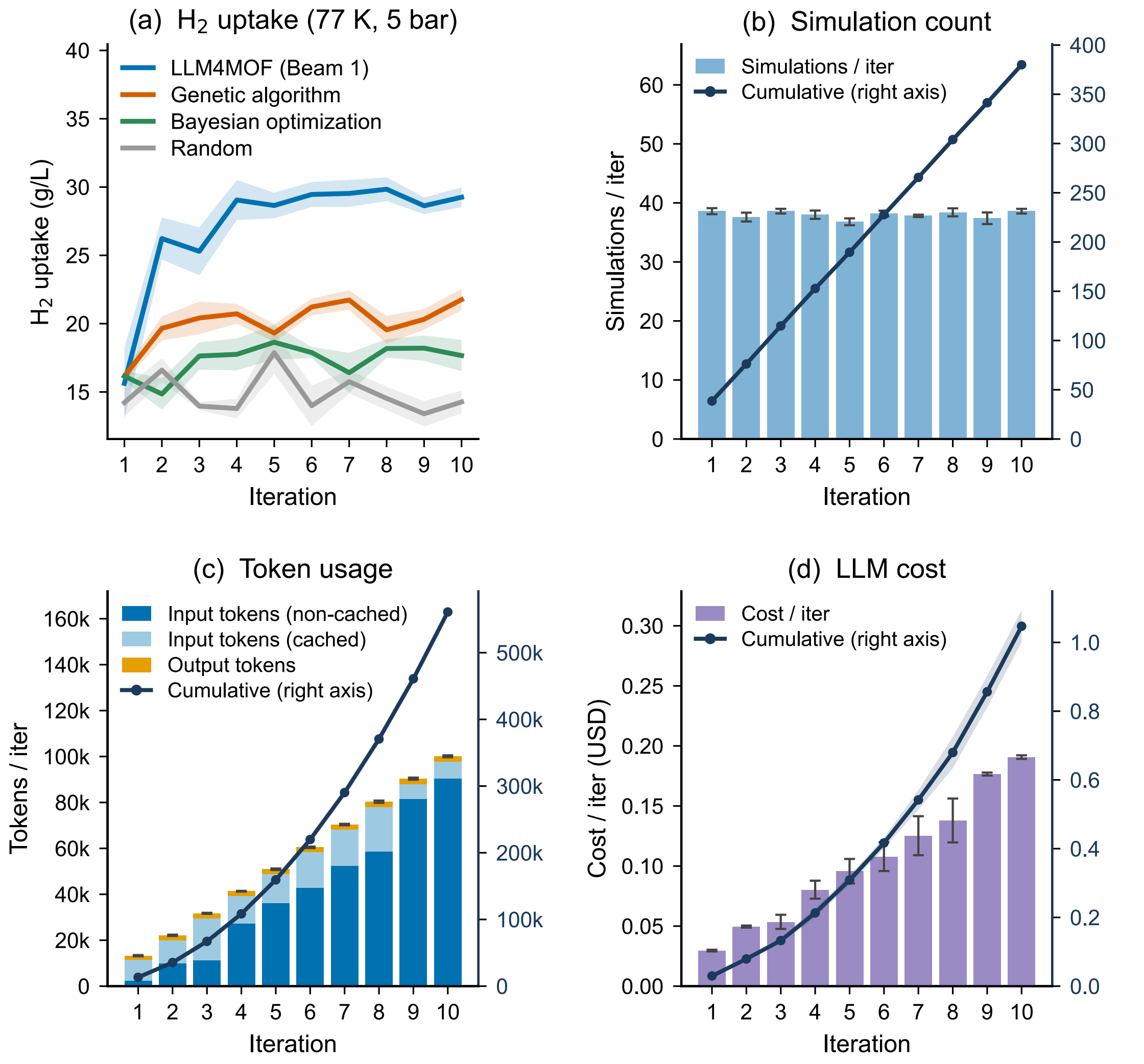

**Figure 6. Efficiency of LLM4MOF in $H_2$ discovery.** (a) $H_2$ uptake at 77 K and 5 bar over ten iterations for LLM4MOF Beam 1, a genetic algorithm[5], Bayesian optimization[23] and random sampling. Lines show, at each iteration, the median across five independent runs of each run's mean uptake, and shaded regions indicate ±1 SEM. (b) Number of RASPA3 GCMC simulations performed per iteration, with the cumulative simulation count shown on the right axis. (c) GPT-5.2 token usage per iteration, separated into non-cached input, cached input, and output

tokens, with cumulative token usage shown on the right axis. (d) LLM inference cost per iteration, with cumulative cost shown on the right axis.

A generative-model baseline is absent from the same-evaluator comparison in **Figure 7a**, as published generative workflows train on thousands to hundreds of thousands of structures.[24,43,44] Even the most label-efficient generative workflow reported requires approximately 1,000 labeled structures, more than double the entire search budget used in this study.[25] To place this cost on the performance axis, that model, which is our own earlier work, was trained at three training-set sizes and evaluated with its published pipeline. With 360 labeled structures, comparable to this work's entire search budget, its generated structures average 27.4 g/L, and only the full 18,463-structure training set raises the mean to 29.4 g/L. Because the two halves of Figure 7a use different simulation codes, these values are reported on their own axis and are not set against the live-loop values. The up-front labeling costs are shown beside the search budgets in **Figure 7b**. Convergence is also fast. In all three discovery domains the loop holds 92 to 98% of its final best by the sixth iteration, although the quoted budget remains the designed 400 evaluations. More generally, the framework is compatible with other expensive property-evaluation modules, including DFT-based band-gap or reaction-energy calculations, as long as the resulting property values can be returned to the Feedback Generator in the same structured format.

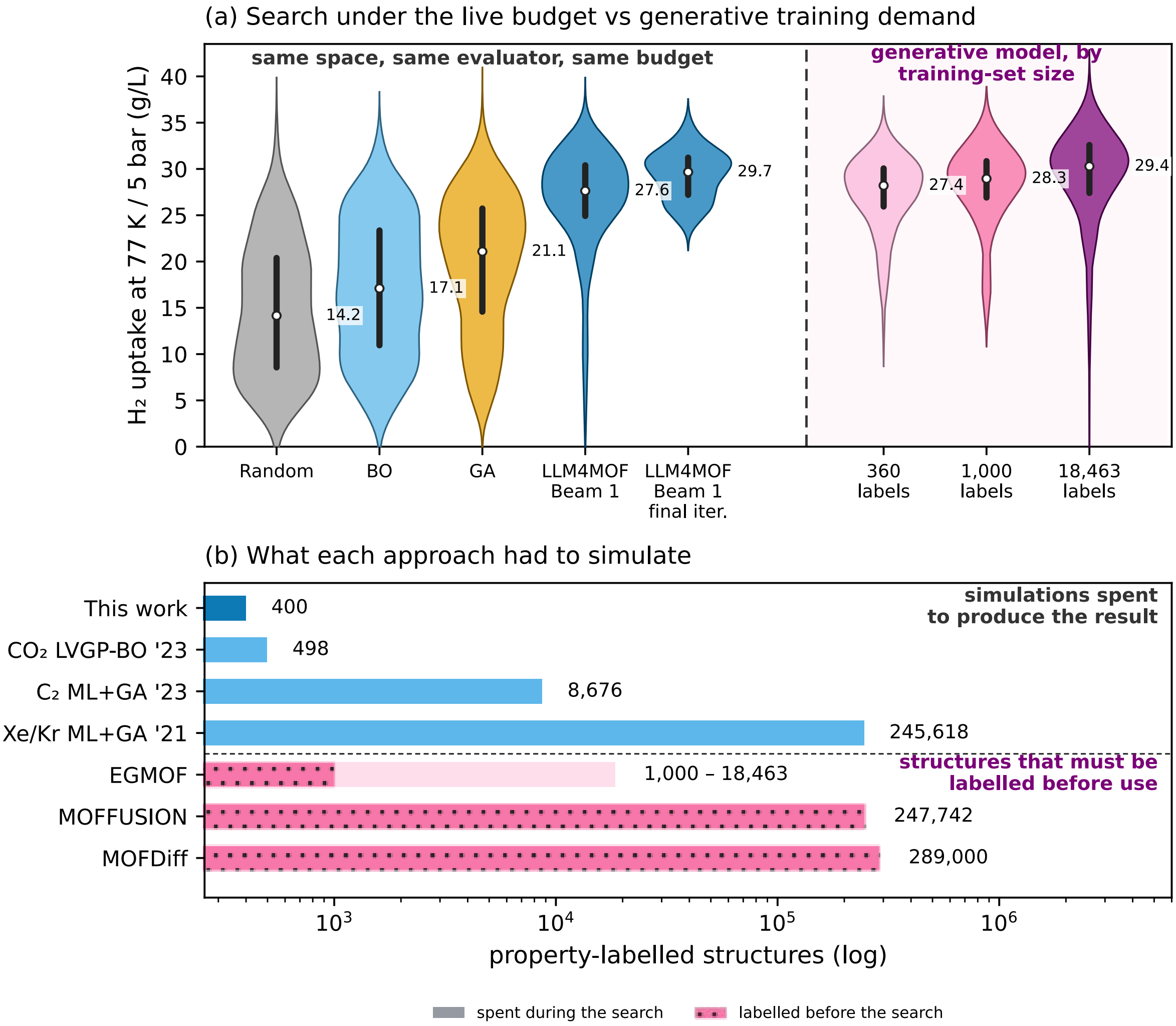

**Figure 7. Comparison with inverse-design approaches.** (a) Per-structure $H_2$ uptake at 77 K and 5 bar. Left of the divider, random sampling, Bayesian optimization, a genetic algorithm, and the full-hypothesis beam search the same design space under the same evaluator and evaluation budget. White dots mark medians and vertical bars interquartile ranges. Right of the divider, a conditional generative model[25] trained on 360, 1,000, and 18,463 labeled structures generates at its best-performing target value and is evaluated with its authors' simulation pipeline. The two halves use different GCMC codes and are not compared numerically. (b) Property-labeled structures consumed by each approach: simulations spent during the search by this work and published campaigns[20,21,23] against training sets required before conditional generation[24,25,43]. A minimum training requirement is reported only for the generative model. Sample sizes and accounting rules are in **Supplementary Note S14**.

# DISCUSSION

We introduced LLM4MOF, a closed-loop framework in which language-model agents propose interpretable design hypotheses, automated modules construct and evaluate the corresponding MOFs, and the results refine the next round of hypotheses. The loop reaches top-performing structures across six adsorption, separation, and electronic-structure tasks within roughly 400 property evaluations, generates and validates new MOFs de novo by live simulation, and does so without training a property-specific model for any objective.

The central contribution is not that a language model can drive MOF search, but that it can do so attributably. Because each round's candidates are organized into diagnostic beams that isolate geometry, chemistry, and metal choice, each outcome reveals which of these was responsible, and this attribution is recovered separately for each property rather than assumed in advance. Such design-level explanation is difficult to extract from surrogate or generative approaches, which return performant structures without the reasoning behind them.

These observations also bear on whether the model reasons or merely orchestrates a constrained search. Which axis dominated was determined per property from the loop's own beam evidence, within a fixed menu of candidate axes supplied to Agent 1 (Note S9). For example in the $CO_2$ task, the loop adopted a textbook open-metal-site prior, found it contradicted by feedback, and revised toward a different mechanism, a correction that non-adaptive search cannot produce; and in discovery mode, run independently of any database, it arrived at the same compact-micropore principle identified in database mode, indicating a genuine, reproducible design rule rather than memorized lookups. The current scope is nonetheless bounded as hypotheses are limited to two-connected linkers and a single objective at a time, and discovery-mode validation covered three adsorption and separation objectives, though the same loop extends directly to other simulation- and DFT-based targets. A like-for-like ranking against conditional generative models is not defined in this setting, because the two families occupy different data regimes. A generative model's labeled training set buys a reusable any-target generator, whereas a search loop's evaluations buy one campaign's outcome. We therefore compare the cost structure and the interface of each family on its own terms, together with a same-evaluator comparison wherever one is possible.

Finally, a more fundamental opportunity lies in the building blocks themselves. In this work, candidates were assembled from PORMAKE's fixed library, but the model reasons over chemical features, not library entries, and a feature description need not match any existing component. In principle, the same loop could propose linkers or metal nodes beyond any current library, thereby expanding rather than merely recombining the design space.

Genetic algorithms are confined to a predefined building-block set, and while generative models can produce genuinely new components, they do so only after training on large structural or property-labeled datasets, the very data that is scarce for most MOF objectives. LLM4MOF instead draws novel chemical proposals from knowledge already encoded in the language model, requiring no task-specific training set. The path forward is to let the loop not only reason about new building blocks but construct them: an inverse design that invents its own components rather than choosing among existing ones.

# METHODS

*LLM-Guided Hypothesis Generation and Constraint Translation*

The LLM agent module in LLM4MOF consists of a hypothesis generator (Agent 1) and a constraint translator (Agent 2). Both agents used GPT-5.2. Agent 1 generates and refines MOF design hypotheses from the user-specified target. At the initial iteration, it receives only the target query; in subsequent iterations, it receives structured feedback from the Feedback Generator while retaining the multi-turn conversation history. Its prompt asks the model to identify the property-governing mechanism, select the dominant design axis among geometry, chemistry, and electronic structure, and propose node and linker features to test the hypothesis. Agent 1 returns a strict JSON hypothesis containing the target application, mechanism, node composition, linker composition, optional pore-geometry window, and iteration-level reasoning. It is not allowed to name topology codes or specific materials, and linker hypotheses are restricted to two-connected linkers.

Agent 2 converts the Agent 1 hypothesis into a constraint JSON without adding new chemical assumptions. Unlike Agent 1, Agent 2 does not retain conversation history and translates only the current Agent 1 hypothesis. It extracts node constraints, linker constraints, global requirements, and an optional geometry filter using controlled-vocabulary tags. Unspecified constraints are left empty or null. The output contains a node query, linker query, global requirements, and optional geometry filter. Representation-specific rules are supplied through a separate rule block, while the core translation prompt is kept unchanged. More detailed descriptions and examples of Agent 1 and Agent 2 are provided in **Supplementary Notes S9 and S10**.

For the backend benchmark, the six Agent 1 backends were accessed through their public APIs with pinned model identifiers, and Agent 2 was pinned to a single model across all arms. Each task-replicate cell is scored on its first attempt: its endpoint is the final-iteration full-hypothesis percentile of the first completed run, and cells whose first attempt did not complete ten iterations are reported as incomplete rather than substituted.

*Feedback Generator*

At each refinement iteration $i$, the feedback message contains the current evidence report in full, while feedback from earlier iterations is compressed into a fixed-size memory ledger. This preserves the most recent evidence verbatim, keeps the prompt length bounded, and limits uncontrolled carryover of earlier observations. The memory ledger is a factual record of prior evidence, not a prescriptive optimization history.

Each feedback report contains up to ten candidates from each of four diagnostic beams: the full hypothesis, chemistry-only sampling with the geometry window removed, metal-only sampling with linker and geometry unconstrained, and random sampling from the full design space. For notation, we refer to these beams as full, chem, metal, and rand. For each candidate, the report includes the target value, chemical profile, and pore geometry under anonymized labels to prevent external lookup. Candidates within each beam are sampled with metal balancing by drawing round-robin across distinct metal identities. Details about the PORMAKE design space can be found in **Table S5**. In database mode the random baseline is drawn by stratified random sampling, balanced across metal identities, so that its reference distribution is not dominated by the database's most common metals. In discovery mode it is drawn uniformly at random from the generative space.

Let $\mathcal{F}_i^b$ denote the feedback from beam $b$ at iteration $i$, and let $\mathcal{F}_i = \bigcup_{b\in\{\text{full,chem,metal,rand}\}} \mathcal{F}_i^b$ be the current four-beam feedback. The memory ledger is constructed only from non-random beams from previous iterations:

$$\mathcal{A}_{<i} = \bigcup_{j=0}^{i-1} \bigcup_{b\in\{\text{full,chem,metal}\}} \mathcal{F}_j^b$$

The user message supplied to Agent 1 is then

$$u_i = \text{Wrapper}[D(\mathcal{A}_{<i}) \parallel \mathcal{F}_i]$$

where $D$ denotes distillation into the fixed-size ledger and $\parallel$ denotes text concatenation.

The random-baseline beam is excluded from $\mathcal{A}_{<i}$ because it is not conditioned on the agent's hypotheses. Including it in the ledger frontier or best-value statistics could confound the refinement signal by mixing hypothesis-derived evidence with full-space random contrast. Nevertheless, it is retained in the current feedback as an unconditioned baseline for assessing whether chemistry-conditioned sampling outperforms random sampling. The ledger records only factual summaries of candidates previously observed by Agent 1, including the best target value, the current top-K frontier, the lower-edge cutoff and median of that frontier, the pore-geometry envelope, and the trajectory of these quantities over earlier iterations. Additional details, an example feedback report, and the memory-ledger ablation study are provided in **Supplementary Notes S3, S11, Table S6 and Figure S7**.

*Simulation Details*

Candidate MOFs were assembled from topology, node, and linker combinations using PORMAKE.[5] The assembled CIF structures were converted to LAMMPS input files using lammps-interface with universal force field (UFF) parameters.[45,46] Geometry optimization was performed in LAMMPS under periodic boundary conditions using conjugate-gradient minimization with alternating fixed-cell atomic minimization and anisotropic cell relaxation. Geometric descriptors of the relaxed structures were recalculated using Zeo++.[47] Although MOF2Zeo was used as a fast geometric pre-filter, all final geometric descriptors were obtained from Zeo++ after LAMMPS relaxation. These descriptors included accessible surface area, unit-cell volume, void fraction, framework density, largest included sphere diameter, largest free sphere diameter, and diffusion-limiting sphere diameter.

Adsorption simulations were carried out using grand canonical Monte Carlo (GCMC) simulations in RASPA3.[48] For hydrogen adsorption, $H_2$ molecules were modeled using a united-atom representation, and the pseudo-Feynman–Hibbs model was applied to account for quantum effects at low temperatures.[49] Framework atoms were described using UFF, cross-interactions were modeled using the Lorentz–Berthelot mixing rule, and a cutoff distance of 12.8 Å was used for van der Waals interactions. For the $C_2H_6$/$C_2H_4$ and $SF_6$ campaigns, ethane and ethylene were described with rigid TraPPE united-atom models[50,51] and $SF_6$ with the Dellis–Samios force field.[52] Details of the simulation parameters are provided in **Supplementary Note S12.**

# Data availability

The source data behind every figure and table, the agent reasoning traces, the structures named in the paper and the held-out split behind Supplementary Figure S6 are available at https://github.com/kn1218/LLM4MOF under source_data, together with the reference property tables, the building-block library and the force fields.

# Code availability

The framework, the agent prompts and the simulation configuration are available at https://github.com/kn1218/LLM4MOF.

## ASSOCIATED CONTENT

**Supplementary Information**. The Supplementary Information is available free of charge and includes detailed descriptions of the work.

## AUTHOR INFORMATION

### Corresponding Author

* Email: jihankim@kaist.ac.kr

Author Contributions

K.N. and S.H. contributed equally to this work: They conceived the research idea, designed and implemented the machine learning model architecture, and conducted the main computational experiments. J.K. supervised the overall project. K.N. and S.H. wrote the manuscript with editorial and discussion inputs from all co-authors. All authors have contributed to the discussions that informed the research and have given approval for the final version of the paper.

**ORCID**

Kyungmin Nam: 0009-0003-8225-1873

Seunghee Han: 0000-0001-8696-6823

Jihan Kim: 0000-0002-3844-8789

**Notes**

The authors declare no competing interests.

**ACKNOWLEDGEMENTS**

This work was supported by the National Research Foundation of Korea (NRF) (RS-2024-00337004 and RS-2024-00451160) and by the InnoCORE program of the Ministry of Science and ICT (N10260141).